\pdfoutput=1
\documentclass[11pt]{article}
\usepackage{acl}
\usepackage{times}
\usepackage{latexsym}
\usepackage[T1]{fontenc}
\usepackage[utf8]{inputenc}
\usepackage{microtype}
\usepackage{graphicx}
\usepackage[normalem]{ulem}
\usepackage{tabularx}
\usepackage{makecell}
\usepackage[colorinlistoftodos]{todonotes}
\usepackage{longtable}
\usepackage{tabularx}
\usepackage{multirow}
\usepackage{fancyvrb}
\usepackage{makecell}
\usepackage{xcolor,colortbl}
\usepackage{comment}
\usepackage{subcaption}

\let\tempone\itemize
\let\temptwo\enditemize
\renewenvironment{itemize}{\tempone\addtolength{\itemsep}{-0.5\baselineskip}}{\temptwo}

\let\tempone\enumerate
\let\temptwo\endenumerate
\renewenvironment{enumerate}{\tempone\addtolength{\itemsep}{-0.2\baselineskip}}{\temptwo}

\title{PriMock57: A Dataset Of Primary Care Mock Consultations}
\author{
  Alex Papadopoulos Korfiatis \\
  Babylon \\
  \texttt{alex.papadopoulos\textsuperscript{1}}
  \And
  Francesco Moramarco \\
  Babylon, University of Aberdeen \\
  \texttt{francesco.moramarco\textsuperscript{1}} \\
  \AND
  Radmila Sarac \\
  \texttt{radmila.sarac@gmail.com} \\
  \And
  Aleksandar Savkov \\
  Babylon \\
  \texttt{sasho.savkov\textsuperscript{1}}
  \AND
  \texttt{\textmd{\textsuperscript{1}@babylonhealth.co.uk}}
}

\begin{document}
\maketitle
\begin{abstract}
Recent advances in Automatic Speech Recognition (ASR) have made it possible to reliably produce automatic transcripts of clinician-patient conversations. However, access to clinical datasets is heavily restricted due to patient privacy, thus slowing down normal research practices. We detail the development of a public access, high quality dataset comprising of 57 mocked primary care consultations, including audio recordings, their manual utterance-level transcriptions, and the associated consultation notes. Our work illustrates how the dataset can be used as a benchmark for conversational medical ASR as well as consultation note generation from transcripts.
\end{abstract}
\section{Introduction} \label{sec:intro}
The use of Automatic Speech Recognition (ASR) is widespread in the clinical domain but it is generally used to alleviate the administrative burden of clinical notes through dictation \citep{hodgson_risks_2016,kumah-crystal_electronic_2018}.

However, the adoption of telemedicine, especially in primary care, generates vast quantities of clinical interaction recordings. Additionally, ASR models have become much more robust to applications in the clinical domain. In turn, this is beneficial for downstream Natural Language Processing (NLP) tasks, such as information extraction from clinical conversations \citep{selvaraj_medication_2021,soltau_understanding_2021} and automatic generation of consultation notes \citep{finley_automated_2018, enarvi_generating_2020, quiroz_identifying_2020, molenaar_medical_2020}.

Despite this being an active area of research it still lacks a commonly recognised ASR benchmark due to the sensitive nature of clinical conversations. Furthermore, as the datasets are not shared, research teams always need to invest time and resources into making their own private dataset. These limitations slow down progress in the field.

We release\footnote{\url{https://github.com/babylonhealth/primock57}} a high quality public dataset of primary care consultation audio recordings, including manual transcriptions and associated consultation notes, which is the basis of our contributions: 
\begin{itemize}
\itemsep-0.2em
\item a benchmark for ASR for primary care conversations;
\item a benchmark for automatic generation of consultation notes for primary care.
\end{itemize}


\begin{figure*}[t]
    \centering
    \includegraphics[width=\linewidth]{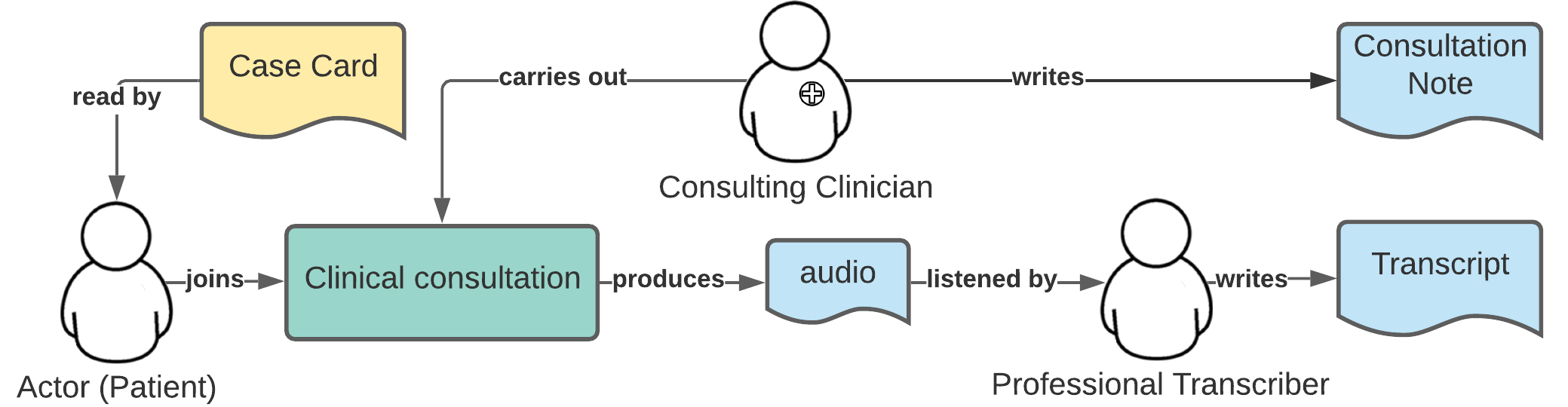}
    \caption{Overview of the data collection process. A mock patient, reading from a medical case card, has a consultation with a clinician which is recorded and transcribed. The resulting dataset includes the consultation audio recordings, notes and manual transcripts.}
    \label{fig:collection-chart}
\end{figure*}

\section{Related Work}

\textbf{Automated transcription of clinical consultations} has attracted quite significant research interest; however, as mentioned above, there is no easily accessible common benchmark dataset in the style of Switchboard \citep{godfrey_switchboard_1992} or Fisher \citep{cieri_fisher_2004}, which are both non-medical conversational audio datasets. Because of this, comparing different approaches for clinical conversation ASR is challenging.

For example, \citet{chiu_speech_2018} detail a dataset of $\approx$ 14,000 hours of recorded and manually transcribed consultations that they use to train an end-to-end clinical conversation ASR model. 
Similarly, \citet{kim_end--end_2020}, \citet{soltau_understanding_2021} develop end-to-end ASR models for clinical conversations and \citet{mani_towards_2020} train a sequence-to-sequence machine translation model to correct the errors of general-domain ASR engines; but they all use different, proprietary datasets.
\citet{johnson_systematic_2014} and \citet{kodish-wachs_systematic_2018} perform systematic reviews of the accuracy of a number of open-source and commercial ASR models for clinical conversation transcription; again, on proprietary datasets.

As for open-access datasets, \citet{he_meddialog_2020} compile and release two clinical dialogue datasets in Chinese and English, covering a wide range of clinical specialties. \citet{ju2020CovidDialog} do the same for COVID-19 related clinical dialogue. These datasets are gathered from online clinical question answering sources; while they are relevant for clinical chatbot research, they are not representative of clinical interactions and do not include audio.
\citet{nazmul2020} provide a dataset of audio recordings, automated transcripts and consultation notes for 70 mock psychiatric consultations --- but no human transcripts.

\textbf{Automatic consultation note generation} and other long-form text summarisation tasks have rapidly developed due to recent advances in Natural Language Generation (NLG) architectures \cite{vaswani2017attention, devlin2019bert}. Several studies \cite{liu2019topic, macavaney2019ontology, zhang2020optimizing, enarvi2020generating, joshi2020dr,krishna-etal-2021-generating, chintagunta2021medically,yim2021towards, moramarco2021preliminary, zhang2021leveraging} use proprietary datasets of transcripts and notes to train NLG models end-to-end, and a number of them carry out automatic or human evaluations on their proprietary test sets.  However, in a similar fashion to the ASR studies discussed above, most studies don't publish these resources; hence, it is again prohibitively difficult to compare their proposed methods. \citet{nazmul2020} provide the only open access clinical dataset that could be used as a benchmark but it only contains psychiatric consultations, which is less applicable to primary care.



\begin{table}[t]
\centering
\begin{tabular}{lr}
\textbf{Consultation type}  & \multicolumn{1}{l}{\textbf{Count}} \\ \hline \hline
Otitis                      & 2                                  \\
Anaphylactic reaction       & 3                                  \\
Cardiovascular              & 11                                 \\
Dermatitis                  & 4                                  \\
Fever                       & 4                                  \\
Urinary tract infection     & 6                                  \\
Upper respiratory infection & 6                                  \\
Asthma                      & 2                                  \\
Gastroenteritis             & 8                                  \\
Mental health               & 3                                  \\
Physical injury             & 2                                  \\
Migraine                    & 6                                 \\ \hline
\end{tabular}
\caption{A breakdown by consultation case card. The case card diagnoses were selected to be representative of common telemedecine presenting complaints.}
\label{tab:consult-types}
\end{table}

\begin{table}[t]
\centering
\begin{tabular}{|p{0.9\linewidth}|}
\hline
\textbf{Demographics (age, gender):} \\
23 year old female \\ \hline
\textbf{Presenting Complaint:} \\
Lower abdominal pain \\
Duration of symptoms: 2 days \\ \hline
\textbf{History, on open questioning:} \\
Have a terrible ache in my lower tummy and feeling hot and sweaty. \\ \hline
\textbf{Symptoms and risk factors:} \\
There is some blood in the urine – pink colour \\
Pain below belly button \\
Feeling nauseated but no vomiting \\
\ \ \ \ \ \ \ \ \ \ \ \ \ \ \ \ \ \ \ \ \ \ \ \ \ \ \ \ \ \ \ * * * \\
\hline
\end{tabular}
\caption{An abridged example of a clinical case card for a Urinary Tract Infection. Mock patients were given a case card and asked to study it before consulting with the clinician. Full version available in the Appendix.}
\label{tab:case-card}
\end{table}

\section{Dataset}
The requirements for releasing a dataset containing Personal Health Information (PHI) are typically costly and involve collecting patient consent and/or de-identification, which is especially challenging with audio recordings. We built a mock consultation dataset as close as possible to the real conditions as a pragmatic alternative. The diagram in Figure \ref{fig:collection-chart} shows an overview of the data collection process.

\subsection{Mock consultation recordings}
We employed 7 clinicians and 57 actors posing as patients from a range of ethnicities. The clinicians had experience with virtual consultations. Participation was optional and anyone could choose to withdraw at any time.
Four of the clinicians were men and three were women; five of them had British English accent, and two of them Indian.
The patient accent distribution is as follows: British English (47.4\%), various European (31.6\%), other English (10.5\%), and other non-English (10.5\%). The gender distribution was relatively even (52.6\% women, 47.4\% men); most participants were from 25 to 45 years old (see Figure \ref{fig:speaker-stats}).

Each mock patient was given a case card that included background information (age, social history, family history of illnesses) as well as information about their presenting complaint, symptoms, conditions, and medications. The case cards were drawn from a pool of primary care conditions, representative of presenting complaints in UK primary care. For a breakdown of presenting complaints, see Table \ref{tab:consult-types}. An example case card is given in Table \ref{tab:case-card}.

We recorded 57 mock consultations (8h38m6s in total) over 5 days, using proprietary telemedicine software that allowed us to export the individual clinician and patient audio channels.\footnote{Due to limitations of the software, audio was exported in compressed form (WebM encoder, Opus codec at a variable bitrate).} In order to emulate real clinical practice, clinicians were using laptops while patients were using mobile phones in an office environment with background noise. Clinicians were asked to act as close as possible to their actual consultation sessions, including conforming to a consultation length of 10 minutes and writing a consultation note in the SOAP format \citep{pearce2016essential}. The resulting mock consultations ranged between 3m48s and 14m18s, with an average consultation length of 9m5s.

\subsection{Manual transcription}
To transcribe the consultation recordings, we employed transcribers with experience in the clinical conversation domain, who were asked to:
\begin{enumerate}
    \itemsep-0.3em
    \item Listen to the consultation audio recordings, in separate channels for clinicians and patients;
    \item Identify the start and end points of individual utterances (continuous speech segments ending in a pause);
    \item Provide an accurate transcription of each of the utterances identified.
\end{enumerate}
Thus we obtained a collection of start times, end times, and utterance-level transcriptions, important for the ASR evaluation described below.

\begin{figure}
    \centering
    \includegraphics[width=\linewidth]{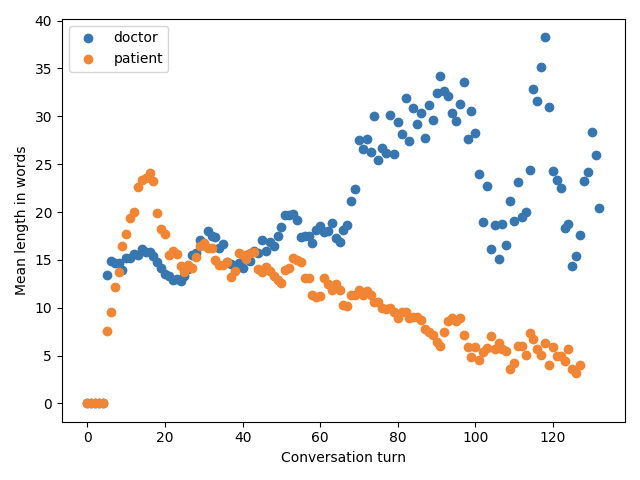}
    \caption{Average utterance length for clinician and patient as a function of conversation turns. The patient initially speaks more than the clinician but later in the consultation this trend is reversed.}
    \label{fig:conversation-turns}
\end{figure}

Consultations have 92 conversation turns and 1,489 words on average; clinicians tend to speak more than patients (897 vs. 592 words per consultation) and take longer turns (19.3 vs 12.8 words per turn).
Interestingly, patients tend to take longer turns than clinicians in the beginning of the consultation, where they presumably state their presenting complaint; turns are more balanced in the middle, and clinicians seem to take over during the diagnosis and management at the end (see Figure \ref{fig:conversation-turns}).

\section{ASR Benchmark} \label{sec:stt}

\begin{table*}[!ht]
\centering
\setlength\tabcolsep{5pt} 

\begin{tabular}{llllllllp{1.1cm}lll}
    & \multicolumn{8}{c}{\textbf{WER}}                                                                                                                                               & \multicolumn{3}{c}{\textbf{ECCA}}                  \\ \cline{2-12}
                & \multicolumn{1}{l}{\textbf{}} & \multicolumn{1}{l}{\textbf{}} & \multicolumn{2}{c}{\textbf{Gender}} & \multicolumn{2}{c}{\textbf{Role}}  & \multicolumn{2}{c}{\textbf{Accent}} & \multicolumn{3}{l}{\textbf{}}                      \\ \cline{2-12}
\textbf{ASR}       & \textbf{mean}                 & \textbf{stdev}                & \textbf{M}       & \textbf{F}       & \textbf{Clinician} & \textbf{Patient} & \textbf{en-gb}   & \textbf{other}   & \textbf{Pr} & \textbf{Re} & \textbf{F1} \\ \hline \hline
GC STT & \textbf{30.9}$\dagger$                          & 12.7                          & 32.7             & 28.9             & 28.5            & 33.4             & 30.0               & 32.2             & 0.83               & \textbf{0.82}            & 0.81        \\
Azure STT  & \textbf{31.3}$\dagger$                          & 12.8                          & 32.7             & 29.6             & 26.7            & 35.8             & 30.2             & 32.7             & \textbf{0.87}               & 0.79            & \textbf{0.82}        \\
ATM & 34.0$\ddagger$                            & 13.9                          & 33.8             & 34.2             & 32.8            & 35.2             & 31.6             & 37.2             & 0.79               & 0.75            & 0.78        \\
Kaldi  & 48.9                          & 14.9                          & 52.7             & 44.6             & 47.0              & 50.8             & 49.5             & 48.2             & 0.64               & 0.69            & 0.68        \\
QuartzNet   & 46.4                             & 15.5                             & 48.4                & 44.1                & 48.1               & 44.7                & 46.6                & 46.1                & 0.67                  & 0.49               & 0.56
\\
Conformer   & 34.4$\ddagger$                             & 14.5                             & 36.8                & 31.7                & 35.6               & 33.2                & 35.0                & 33.7                & 0.79                  & 0.71               & 0.75
\\ \hline
\end{tabular}
\caption{Word Error Rate (WER) scores for a number of Speech-to-text engines, and Extracted Clinical Concepts Accuracy (ECCA) based on recognised clinical terms. The gender, role and accent breakdowns show how each factor affects the mean WER. $\dagger$ indicates lack of statistical significance between mean WER scores ($p=0.097$); $\ddagger$ is weak significance ($p=0.026$); all other scores are $p<0.001$.}
\label{tab:wer}
\end{table*}


We perform a baseline study of ASR for clinical conversations by passing the audio recordings of the mock consultations through commonly used open-source and commercial speech-to-text engines:
\begin{enumerate}
    \itemsep-0.3em
    \item \textbf{Kaldi}: This is our baseline system, built using the Kaldi \citep{povey2011kaldi} speech recognition toolkit, running locally. It uses a pretrained acoustic model from Zamia Speech\footnote{\url{http://zamia-speech.org/asr/}} and a 3-gram language model trained on a proprietary medical question answering dataset.
    \item \textbf{NeMo QuartzNet \& Conformer}: These systems use QuartzNet \citep{kriman2020quartznet} and Conformer \citep{gulati2020conformer} ASR models, which we load using Nvidia's NeMo toolkit.\footnote{\url{https://github.com/NVIDIA/NeMo}} Both models are end-to-end and do not use a language model.
    \item \textbf{Google Cloud Speech-to-text (GCSTT)}:\footnote{\url{https://cloud.google.com/speech-to-text}} a commercially available, general domain service. We use the \textit{video} enhanced model which is only available for the \textit{en-us} language.
    \item \textbf{Amazon Transcribe Medical (ATM)}:\footnote{\url{https://aws.amazon.com/transcribe/medical/}} a commercially available service, tailored specifically for medical use cases. There are models available for \textit{clinical dictation} and \textit{clinical conversation}; we use the conversation model with \textit{speciality=Primary Care}.
    \item \textbf{Azure Speech-to-text (ASTT)}:\footnote{\url{https://azure.microsoft.com/en-us/services/cognitive-services/speech-to-text/}} a commercially available, general domain service. We use the \textit{Standard} model. 
\end{enumerate}

To test the accuracy of the above services, we first extract the audio for each individual utterance identified by our human transcribers. We then generate a transcript for the utterance using each of the ASR engines. We ensure consistency by performing the following post-processing steps on both human and automatic transcripts:
\begin{itemize}
    \itemsep-0.3em
    \item Remove disfluencies ("umm", "uhh", etc.). These are included in the reference transcripts, but often omitted in each STT service;
    \item Replace numerals ("5", "9th", "1984") with written equivalents ("five", "ninth", "nineteen eighty-four") to ensure uniformity;
    \item Remove all punctuation, collapse multiple spaces and convert to lowercase.
\end{itemize}
Finally, we compute the Word Error Rate (WER) for each utterance using SCTK's \textit{sclite}\footnote{\url{https://github.com/usnistgov/SCTK}} tool. The mean WER, including a breakdown by gender, role, and accent can be seen in Table \ref{tab:wer}. Even though both are general domain, Google and Azure together are the best performing models on our dataset ($p = 0.097$). Conformer performs surprisingly well, given that it is a character-level model evaluated on a word-level metric.

The base WER metric treats all words in a transcript as equally important; this may be less desirable in the clinical domain, where the correct transcription of specific clinical terms is expected to be more important. To test this, we use a proprietary clinical information extraction engine based on fuzzy string matching, linking to SNOMED-CT \citep{donnelly2006snomed}. We extract medical concepts from each utterance in both reference and hypothesis transcripts, then compare the concepts extracted to estimate accuracy based on clinical terminology (ECCA in Table \ref{tab:wer}). The results mostly match the WER comparisons; the medical-domain Amazon model does not seem to perform better.

\section{Consultation Note Generation Benchmark}


%

\label{sec:note-gen}
\renewcommand{\cellalign}{l}
\begin{table*}[!h]
    \setlength{\tabcolsep}{4pt} 
    \def\arraystretch{1.3}
    \centering
    \begin{tabular}{l|p{8.5cm}|p{5cm}}
        \multicolumn{2}{c|}{\cellcolor{blue!25}\textbf{Transcript}} & \multicolumn{1}{c}{\cellcolor{blue!25}\textbf{Note}} \\\hline
          \cellcolor{blue!5}Clinician & \cellcolor{blue!5}So, um, tell me what's been going on. You've been saying there's a problem with your hearing. Is that right? & \multirow{8}{*}{\makecell{History:\\
          Hx of difficulty hearing left ear\\
          for 6 weeks with tinnitus and\\
          slight nausea/ dizziness.\\
          One previous similar episode in\\ the past- resolved spontaneously.\\
          No discharge/fever/itchiness/pain\\
          Doesn't use cotton wool buds\\
          No Pmhx of note\\
          Ex: Looks well, not in pain.\\
          Imp: need to exclude impacted\\ wax in ear canal first\\
          Pln: for face to face GP\\ appointment in 5 days to examine\\
          ear\\
          If any problems in interim to\\ ring us back\\
          Pt happy with and understands\\plan}}\\

          \cline{1-2}
         
         Patient & Yeah, so I just feel I can't really hear as well as I used to, like my hearing is kind of deteriorating in some way. \\\cline{1-2}
         \cellcolor{blue!5}Clinician & \cellcolor{blue!5}Right, OK. How long has this been going on for? \\\cline{1-2}
         Patient & Uh about six weeks. \\\cline{1-2}
         \cellcolor{blue!5}Clinician & \cellcolor{blue!5}Six weeks, OK. Um, and before that have you had any hearing problem at all? \\\cline{1-2}
         Patient & Um I had something maybe, about a year ago, but it only lasted a couple of days, it wasn't anything as long as this.\\\cline{1-2}
         \cellcolor{blue!5}Clinician & \cellcolor{blue!5}Right, OK, OK. And, um, in this six week period, have you had anything else happen? Have you had any other ear symptoms at all? \\\cline{1-2}
         Patient & Um, I occasionally get like a ringing in my left ear, uh just on the one side and um there's actually been a few times when I felt kind of a bit sick or a bit dizzy as well.\\
    \end{tabular}
    \caption{Snippet of a mock consultation transcript and the corresponding note, written by the consulting clinician.}
    \label{tab:transcript-note}
\end{table*}

\begin{table}[!h]
    \centering
    \begin{tabular}{lllll}
        \textbf{Model} & \textbf{R1} & \textbf{R2} & \textbf{RL} & \textbf{B}\\\hline \hline
        BART-CNN & 0.17 & 0.02 & 0.10 & 0.80 \\
        BERT-ext & 0.21 & 0.03 & 0.10 & 0.78 \\
        Random & 0.19 & 0.02 & 0.09 & 0.78\\
        BART-finet & \textbf{0.31} & \textbf{0.08} & \textbf{0.17} & \textbf{0.81} \\ \hline
    \end{tabular}
    \caption{Average common metrics scores of different models on the 57 consultations. R1 through L represent Rouge F1 scores for unigrams, bigrams, and longest-common-subsequence. B represents non-rescaled BERTScore; score range is between 0.7 to 0.9, so differences are less pronounced.}
    \label{tab:note-gen}
\end{table}

The consultation transcripts and corresponding notes (see example in Table \ref{tab:transcript-note}) are intended as a parallel dataset to evaluate methods for automatically generating primary care consultation notes.
We propose a benchmark for this task by evaluating a number of baseline approaches and reporting common automatic metric scores on our dataset. The approaches considered include:
\begin{description}
\itemsep-0.3em
    \item \textbf{BART-CNN}: a neural sequence-to-sequence summariser based on the BART model \cite{lewis-etal-2020-bart} and fine-tuned on the Dailymail/CNN dataset \cite{nallapati2016abstractive};
    \item \textbf{BERT-ext}: a general-purpose extractive summariser based on Bert embeddings \cite{miller2019leveraging};
    \item \textbf{Random}: a baseline that extracts 15 random sentences from the transcript and collates them to form a note;
    \item \textbf{BART-finet}: a BART-CNN model further fine-tuned on a proprietary dataset of 8,000 real transcripts and consultation notes.
\end{description}

We evaluate the models on our dataset and report common summarisation metrics scores: Rouge-1, -2 \& -L \citep{lin2004rouge} which compute the F-score across ngrams between generated and human notes; and BERTScore \citep{zhang2019bertscore}, which computes the similarity between BERT embeddings of the notes.

The results can be seen in Table \ref{tab:note-gen}: the fine-tuned BART model scores highest with all metrics, while \emph{BART-CNN} and \emph{BERT-ext} fail to outperform the \emph{Random} baseline model. This highlights the differences between  consultation note generation and general-purpose summarisation.

A more detailed evaluation of this task can be found in \citet{human-eval}; example notes can be found in Appendix Table \ref{tab:generated-notes}.

\section{Conclusion}
We present a dataset of 57 high quality mocked consultation audio recordings, their manually aligned and diarised transcripts, and consultation notes. By publishing this dataset, we hope to offer a benchmark for future studies in both ASR for clinical conversations and Consultation Note Generation for the primary care domain.

\bibliography{bibliography, custom}
\bibliographystyle{acl_natbib}
\onecolumn
\renewcommand\thefigure{A.\arabic{figure}} 
\renewcommand\thetable{A.\arabic{table}} 
\section*{Appendix}
\label{sec:appendix}
\setcounter{figure}{0} 
\setcounter{table}{0} 

\begin{figure}[!h]
    \centering
    \includegraphics[width=0.49\linewidth]{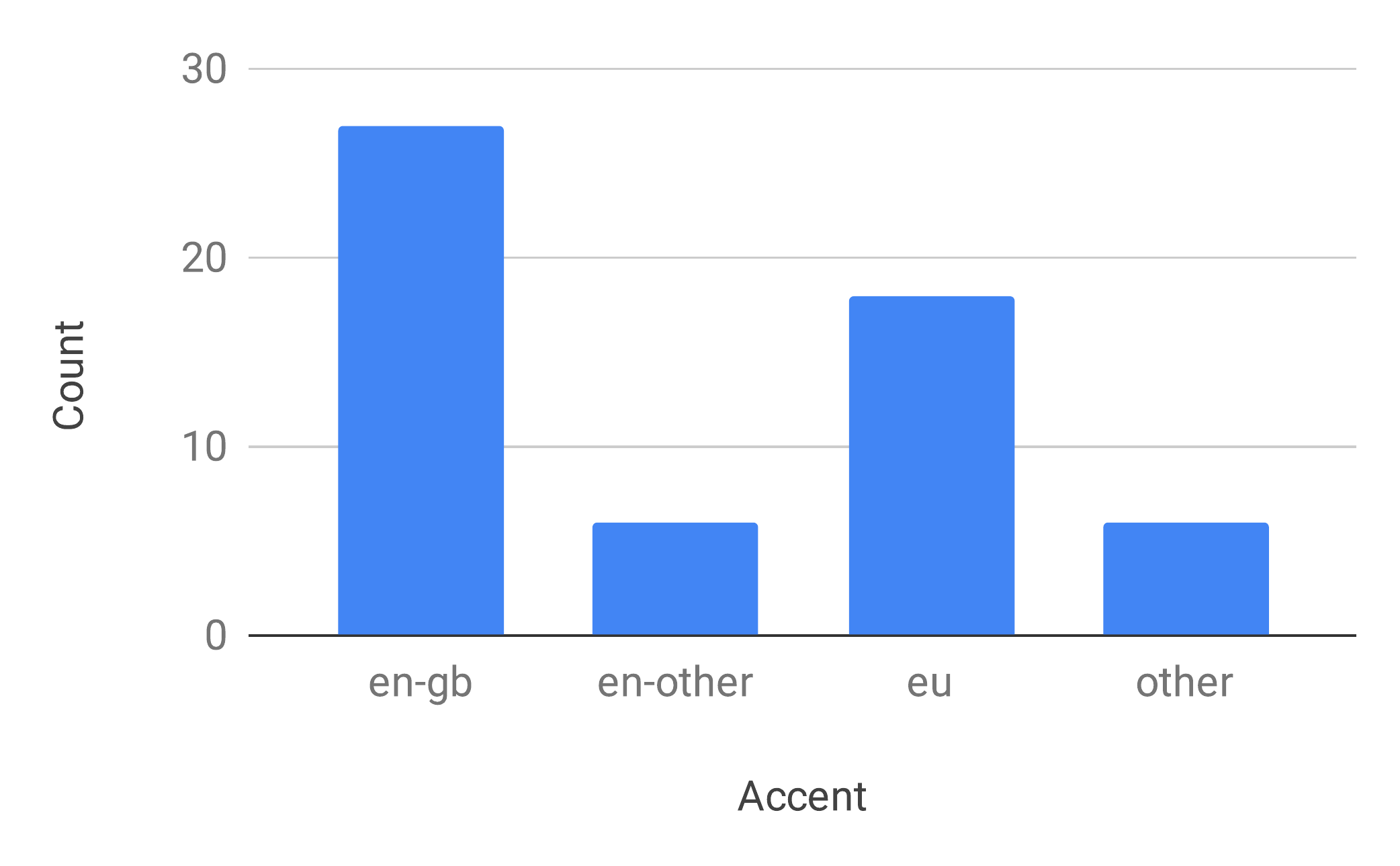}
    \includegraphics[width=0.49\linewidth]{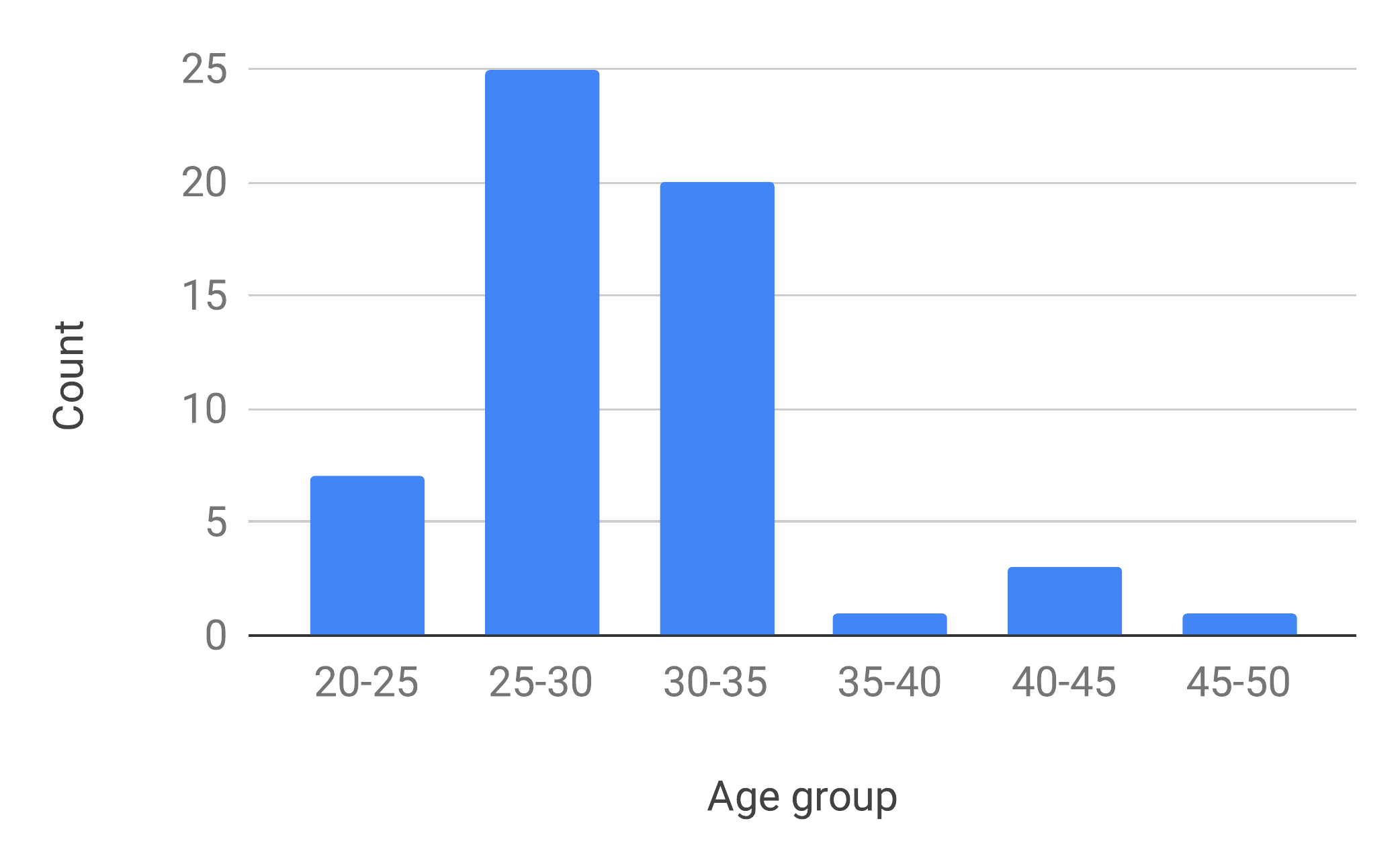}
    \caption{Accent and age group distributions for patients in the 57 mock consultations.}
    \label{fig:speaker-stats}
\end{figure}

\begin{table}[!ht]
\centering
\begin{tabular}{|p{0.9\linewidth}|}
\hline
\textbf{Demographics (age, gender):} \\ \hline
23 year old female \\ \hline
\textbf{Presenting Complaint:} \\
Lower abdominal pain \\
Duration of symptoms: 2 days \\ \hline
\textbf{History, on open questioning:} \\
Have a terrible ache in my lower tummy and feeling hot and sweaty. \\ \hline
\textbf{Symptoms and risk factors:} \\
There is some blood in the urine – pink colour \\
Pain below belly button \\
Feeling nauseated but no vomiting \\
Going to the toilet a little more often but drinking lots of fluids \\
No urine urgency or pain when passing urine. \\
Was constipated until 1 week ago but that has cleared up now \\
Had sexual intercourse 4 days ago \\
No new sexual partner since last STI screen 6 months ago \\
No vaginal discharge \\
Has Implanon contraceptive implant for 1 year \\
No change in vaginal bleeding \\
No loin pain \\
Activities of daily living: No problems performing daily activities \\
Family history: nil \\
Past Medical History: nil \\
Drug History: Implanon \\
Allergies: Amoxicillin \\ 
\hline
\end{tabular}
\caption{Example clinical case card for a Urinary Tract Infection. Mock patients were given a case card and asked to study it before consulting with the clinician.}
\label{tab:case-card-full}
\end{table}

\begin{table*}[t]
    \centering
    \begin{tabular}{|p{7.2cm}|p{7.8cm}|}
    \hline
        \textbf{Human Transcription} & \textbf{Google Speech-to-text} \\ \hline  \hline
\textbf{Doctor:} Hello? \newline
\textbf{Patient:} Hello. Can you hear me well? \newline
\textbf{Doctor:} Uh uh yes. I think. It's a bit better. It's a bit, it's a bit, it's not very clear. But let's continue anyway. \newline
\textbf{Patient:} OK. \newline
\textbf{Doctor:} Uh, OK. Let's start again. So how can I help you sir? \newline
\textbf{Patient:} Yes. So, it's been a few days now. I have like a sore, and a red skin. It's kind of, it's really itchy, and it's like super annoying. So I'd like to find something quick to solve it. \newline
\textbf{Doctor:} OK. No, no problem. I'm happy to help. Um whereabouts in your skin is it affected? \newline
\textbf{Patient:} Uh, mostly like my chest, my, my hands, my arms. Like, like really, it's it's super annoying. Like it's itching a lot, like all the time. And I can't even sleep at night. I really need something quickly to, to solve it. Because even at work I, I can, when I'm in a meeting and I have to, like uh think about my work, I can't focus, I can't actually focus on my work. It's really annoying because I can't actually think about, uh, what I have to say. I'm always like, uh, disturbed by this disease.
\begin{center}
    * * *
\end{center}
\textbf{Doctor:} OK. OK. So it's something for you to think about. you can get different types of antihistamines. I can give you something a little bit stronger today as well. Um, something like Fexofenadine, which I can give to you today. It's definitely worth trying, and it's not going to do you any harm. \newline
\textbf{Patient:} OK. \newline
\textbf{Doctor:} Um but I think using the steroids and the emollients, um on a regular basis Uh over the next week to ten days, should hopefully control your symptoms. But do come back and see me next week, if things don't get better. \newline
\textbf{Patient:} That sounds good. \newline
\textbf{Doctor:} OK? Um do you have any questions for me? \newline
\textbf{Patient:} Uh, no that's it. Thank you very much. Bye. Thank you as well. Bye. \newline

 & 
\textbf{Doctor:} Hello.  \newline
\textbf{Patient:} Hello, can you hear me wet?  \newline
\textbf{Doctor:} Yes, I think it's a bit better. It's a bit. It's a bit. It's not very clear. But let's continue. Anyway,  \newline
\textbf{Patient:} Okay.  \newline
\textbf{Doctor:} okay, let's talk again. So, how can I help you, sir?  \newline
\textbf{Patient:} Yes, so it's been a few days now. I have like a sore and the Redskin it's kind of it's really itchy and it's like super annoying.  \newline
\textbf{Doctor:} Okay.  \newline
\textbf{Patient:} So I'd like to find something quick to serve it.  \newline
\textbf{Doctor:} No, no problem. Happy to help whereabouts of your skin is affected.  \newline
\textbf{Patient:} Mostly like my chest my my hands my arms like agree. It's super annoying like it's itching a lot like all the time and I can't even sleep at night. Like I really need something quickly to study because even at work I like when I'm in the meeting and I have to like think about my work Focus like actually focus on my work. It's  \newline
\textbf{Doctor:} Yeah.  \newline
\textbf{Patient:} really annoying because I can actually think about what happened say, I'm always like disturbed by this disease.
\begin{center}
    * * *
\end{center}
\textbf{Doctor:} It didn't okay. So something for you to think about a you can get different types of and system means I can give you something Little Bit Stronger today as well  \newline
\textbf{Patient:} Okay.  \newline
\textbf{Doctor:} something like fix the penalty in which I can give to you today. It's definitely worth trying it's not gonna do you any harm but I say anything using the steroids and the emollients on a regular basis over the next week to 10 days should hopefully care control your symptoms, but do come back and see me next week if things don't get better.  \newline
\textbf{Patient:} That sounds good.  \newline
\textbf{Doctor:} Okay any questions for me?  \newline
\textbf{Patient:} And now that's it.  \newline
\textbf{Doctor:} Okay. Well, I wish you all the best.  \newline
\textbf{Patient:} Thank you very much.  \newline
\textbf{Doctor:} Hope you have a good day.  \newline
\textbf{Patient:} Bye-bye. \\ \hline
    \end{tabular}
    \caption{An example of a human transcript and a Google Speech-to-text transcript for one of the mock consultations. Both are partial.}
    \label{tab:stt-example}
\end{table*}


\begin{table*}[t]
    \centering
    \begin{tabular}{|l|p{15cm}|}
    \hline
         \rotatebox{270}{\textbf{Human Note}} &
Hx: 1 week history of spontaneous elbow swelling left.
Not painful.
No trauma.
No FH of rheumatological disease- NB pt says he has been old he has OA previously by doctors- ?
need to confirm this
 Works in a desk job
 Not happened before
 Otherwise well- 
 PMHx: nil of note
 FH: nil of note
 DH: not on any medication, allergic to peanuts
 SH: exercises regularly, active
 Ex: looks well, not in pain.
Mild erythema and minimal swelling (if any) around olecranon process left elbow
 Imp: possible bursitis
 Plan: for NSAIDs- usual advice re SE
 For rheum bloods: esr, crp, fbc, rheum factor and urate
 Review thereafter in person/ via video
 To contact us back in interim if any deterioration/concerns- pt warned re symptoms of septic arthritis.\\\hline\hline

\rotatebox{270}{\textbf{{BART-CNN\ \ }}}	&
Doctor Deen Mirza from GP at Hand sees John Smith.
John says he has a weird swelling on his left elbow.
He also says he is allergic to peanuts.
Deen takes a look at John's elbow to see if there is anything wrong with it.
Do you have any other illnesses at all?\\ \hline

\rotatebox{270}{\textbf{BERT-ext}} &  
Before we start your appointment, could you please tell me your first name and your date of birth.
And I was born on the fifth of April, , nineteen seventy three.
But it's just, just a bit, a bit weird, to see that.
, and , , in terms of your job, do you do anything physical?
so you know you said you think you've got , , osteoarthritis.
and, do you have any other illnesses at all?
, I run regularly, like two, three times a week.
, what I think we should do is, I think you should be on some anti-inflammatory medication, in the, in the first instance.
And, there'll be instructions within that pack, about where to go to get those blood tests done.
and , your, your joint doesn't look like that.
However, if your, the elbow was to become very red, very painful, , and the redness was to spread or become , you know more intense.
That would require more immediate assessment, more immediate treatment.
do you, do you think it's something dangerous?
Like something, like could I die from that, or is it, is it No.
that's four hundred milligrams, two times a day.
Maybe within a , actually you know, the follow-up appointment doesn't have to be face-to-face, if it's more convenient for you do, to do it over the phone, we can do that over the phone, , over video.
We can do that as well, that's, that's your call.\\\hline

\rotatebox{270}{\textbf{Random}} &  
Sure.
No, no I haven't noticed that before.
OK, OK, great.
Yes, a few years ago.
do you, do you think it's something dangerous?
Fantastic.
But you contact us, , after you've had the blood test done, and we can review things then, OK. OK. OK, yeah that sounds good.
OK. -.
, yeah, no, I'm, think I'm healthy.
.
So, , this, this is not the case right now.
I run regularly, like two, three times a week.
don't need to worry.
All right then, OK. , take care then.\\\hline

\rotatebox{270}{\textbf{BART-finet}} &
You have a problem with your left elbow.
1 week ago noticed a weird swelling on the left elbow.
Not painful at all, but slightly warm, slightly warm.
No pain, no swelling, no fluid in the elbow.
No injury.
No previous history of this.
No injury to the elbow.
NKDA.
SH: Mobile and active, exercise 2-3 times a week, running.
Osteoarthritis of the elbow.
You should start the treatment you have been prescribed.
You should begin the treatment prescribed as we discussed.
You may want to take some ibuprofen or paracetamol in addition to any prescribed medication.
\\ \hline
    \end{tabular}
    \caption{Examples of a human written note and automatically generated notes with the four baseline models.}
    \label{tab:generated-notes}
\end{table*}
\end{document}